\documentclass{article}
\usepackage{spconf,amsmath,graphicx}
\usepackage{algorithm}     
\usepackage{algpseudocode}
\usepackage{amssymb}
\usepackage{svg}
\usepackage{amsmath}
\usepackage{booktabs}

\title{LLM-augmented symbolic RL with landmark-based task decomposition}
\name{Alireza Kheirandish, Duo Xu, Faramarz Fekri}
\address{School of Electrical and Computer Engineering, 
Georgia Institute of Technology}
\begin{document}
%
\maketitle
\begin{abstract}
One of the fundamental challenges in reinforcement learning RL is to take a complex task and be able to decompose it to subtasks that are simpler for the RL agent to learn. In this paper, we report on our work that would identify subtasks by using some given positive and negative trajectories for solving the complex task. We assume that the states are represented by first-order predicate logic using which we devise a novel algorithm to identify the subtasks. Then we employ a Large Language Model (LLM) to generate first-order logic rule templates for achieving each subtask. Such rules were then further fined tuned to a rule-based policy via an Inductive Logic Programming (ILP)-based RL agent.
Through experiments, we verify the accuracy of our algorithm in detecting subtasks which successfully detect all of the subtasks correctly. We also investigated the quality of the common-sense rules produced by the language model to achieve the subtasks. Our experiments show that our LLM-guided rule template generation can produce rules that are necessary for solving a subtask, which leads to solving complex tasks with fewer assumptions about predefined first-order logic predicates of the environment.


\end{abstract}
\begin{keywords}
Reinforcement Learning, Large Language Model, Inductive Logic Programming, Contrastive Learning
\end{keywords}
\vspace{-0.5cm}
\section{Introduction}
\label{sec:intro}


In the realm of Reinforcement Learning (RL), strategically using landmarks and subtasks is a key technique for managing complex tasks \cite{barto2003recent}. This method systematically breaks down daunting challenges into smaller, achievable goals and clear pathways, making intricate tasks more manageable \cite{pateria2021hierarchical}. To complete a complex task, we must visit certain specific states—referred to as landmarks—that contain essential information for successfully accomplishing the task.
Landmarks act as critical milestones that facilitate effective decision-making and enhance structured, efficient problem-solving strategies \cite{porteous2014extraction}. These landmarks constitute essential milestones about the task, crucial for achieving the goal. For example, a landmark could be possessing specific combinations of objects, arriving at a particular location, or visiting certain places in a specific order \cite{elkawkagy2012improving}.
We define each of these landmarks that are necessary to complete a task as a subtask. Subtasks can consist of either the entire state or a subset of the state.
Subtasks are particularly valuable in complex environments where a straightforward trajectory to the goal is not readily apparent or where the policy required to solve intricate tasks is complex, making straightforward solutions challenging. 

While other works have addressed identifying landmarks through reward-centric algorithms \cite{jothimurugan2023robust,icarte2018using}, our algorithm uses state trajectories labeled only with a single indicator of whether the trajectory was successful in completing the task. This approach is crucial in environments with sparse and non-interpretable rewards. For this purpose, we have used contrastive learning \cite{le2020contrastive} with the logic-predicate representation of the states as its input.

Recently, there has been significant interest in symbolic RL in general \cite{landajuela2021discovering,basu2024explorer}. 
Symbolic RL has the advantage of being human interpretable and also more generalizable to new environments. In particular, as a special type of symbolic RL, inductive logic programming (ILP)-based RL agents \cite{jiang2019neural, payani2020incorporating, xu2022integrating} have utilized differentiable rule learners known as $\partial$ILP \cite{evans2018learning,payani2019inductive} to form logic-based policies. 

Recently, an RL method denoted as NUDGE \cite{delfosse2024interpretable} was proposed using ILP to generate interpretable policy as a set of weighted rules. We will be using NUDGE framework as the ILP engine for further fine tuning our rules generated by LLM for the subtasks.


When processing an input state, the NUDGE system identifies entities and their interactions, transforming raw states into logical representations. In the realm of first-order logic, a predicate functions as a Boolean operation on terms, which are defined as objects or variables. We establish our subtasks using distinct combinations of predicates, thereby facilitating the creation of interpretable subtasks. Our empirical findings indicate that creating subtasks does not require detailed predicates from the environment.


The advent and evolution of Large Language Models (LLMs) have sparked significant interest due to their ability to utilize common sense knowledge and process information in natural language, mirroring real-world understanding \cite{zhang2024llm}. 
There are recent research works that elaborate LLMs as either auxiliary supports or principal agents within RL frameworks \cite{place2023adaptive, tan2024true}. These innovative approaches utilize the descriptive and inferential strengths of LLMs to more effectively navigate and solve complex environmental challenges. By synergizing LLMs' linguistic capabilities, they push the boundaries of what intelligent systems can achieve in tackling complex tasks \cite{ahn2022can}.

The generation of related rules represents the initial step for an ILP-based RL agent to establish rule-based policies. Generating a comprehensive rule space from scratch in symbolic RL presents significant challenges due to the vastness of the potential rule space \cite{jiang2019neural}. Previous works have addressed this problem by using algorithms based on human expert rule templates \cite{delfosse2024interpretable}. However, our approach leverages the common sense knowledge embedded within LLMs to efficiently generate the necessary rules. We replace human-generated rule templates with LLM-generated rule templates, empirically demonstrating that our approach is as efficient as other rule generation algorithms.\footnote{{https://github.com/KheirAli/LLM\_Landmark.git}}



In section 2, we introduce our algorithm for identifying necessary subtasks.
Section 3 presents the LLM rule generation technique. In this section, we explore how we can utilize the interpretable subtasks generated from the previous section to develop further interpretable rules. These rules are then used as rule templates for an ILP-based RL agent, formulating a rule-based policy.

\vspace{-0.3cm}
\section{Landmark Identification from trajectories}
\label{sec:format}

Reinforcement learning (RL) tackles decision-making problems in environments defined by a state space \( S \), an action space \( A \), and transition dynamics \( P(s' \mid s, a) \), where the goal is to maximize rewards over time. In this context, a policy \( \pi(a \mid s) \) maps states to actions, guiding the agent towards maximizing the expected discounted sum of rewards 
$E_{\pi} \left[ \sum_{t} \gamma^t r(s_t, a_t, s_{t+1}) \right]$,
where \( \gamma \) is a discount factor to prioritize immediate rewards. This sets the stage for designing RL algorithms that can learn optimal actions in complex decision spaces.

Incorporating a rule-based policy within this RL framework can provide a structured and interpretable way to guide decision-making. Leveraging concepts from First-Order Logic (FOL), we represent policies as rules. In FOL, predicates describe relationships between terms (constants, variables, or function-based expressions), \( p(t_1, \dots, t_n) \), and rules consist of a head (the action to be taken) and a body (a set of predicates describing the current state). Rules are often written in the form \( A :- B_1, \dots, B_n \), where \( A \) is the head (action) and \( B_1, \dots, B_n \) are the body predicates.

In our approach, we employ an ILP-based RL agent, as described in the NUDGE \cite{delfosse2024interpretable}, with states represented by grounded FOL predicates. To identify landmarks, we first apply a contrastive learning algorithm to detect potential landmark states, followed by a graph search algorithm \cite{muggleton1991inductive} to identify the necessary grounded predicates for each subtask. We leverage both positive and negative trajectories from a Neural Network (NN) RL agent, collecting 50 positive and 500 negative trajectories during the early stage of training. The advantage of using an NN agent is that it does not require prior information about the environment. Positive trajectories are those that successfully achieve the task's goal, while negative ones do not.

\sloppy
Each state trajectory is defined as $\tau_i$, where $\tau_i = (s_0, s_1, \ldots, s_T)$.
$\tau^p_{i}$ is the i'th positive trajectory and $\tau^n_{i}$ refers to i'th negative trajectory. We used a two-layer NN to assign a number between zero and one to every state. We propose that landmarks should consistently appear in all positive trajectories but may occasionally appear in some negative ones. To achieve this, we train the NN to output 1 for landmark states and 0 for non-landmark states. For this aim, we should maximize this function:

\[
\sum_{(\tau^p_{i},\tau^n_{j})}\log(\frac{\exp{(\sum_{s_k}f_\theta(\tau^p_{i}(s_k)))}}{\exp{(\sum_{s_k}f_\theta(\tau^p_{i}(s_k)))}+\exp{(\sum_{s_k}f_\theta(\tau^n_{j}(s_k)))}})
\]

where $\tau^p_{i}(s_k)$ denotes the k'th state of i'th trajectory of positive samples. The sum is over the pairs of randomly chosen trajectories from positive and negative samples. The results of the algorithm are detailed in the experimental section of this paper.

Next, we develop a method for identifying subtasks from our landmark candidates. The necessity of subtasks in every positive trajectory is a characteristic that stems from the definition of a subtask. A subtask is defined as a necessary state or subset of a state that must be visited to complete a task.

The algorithm takes as its input the set of all candidates' landmark states resulting from the contrastive learning algorithm. Then it proceeds to evaluate all combinations of grounded predicates to identify all subtasks. 
As shown in Fig. 1, we associate all of the predicates to $Node_{00}$ at the root of the tree graph.
A subtask is defined by its consistent presence in every positive trajectory and its absence in negative trajectories, which we verify by examining random negative samples. If no subtask is detected at the current node, we extend the tree graph by adding leaves. Each leaf is created by removing a predicate from the current set assigned to the node, move to a deeper level, and add the newly formed nodes to the frontier.


To determine the next node to explore from the frontier, a softmax function is applied on $f(Node)$, which is based on two factors: the number of unique predicate combinations in the node and its level in the search hierarchy. Our goal is to find the largest set of predicates that define a subtask. Once a node is validated, it is explored further by increasing its level and removing it from the frontier. Details are provided in Algorithm 1.

Our graph search algorithm identifies the largest set of predicates that reliably activate landmarks, treated as subtasks for the next stage. Fig. 2 highlights how the graph search enhances the algorithm's precision and efficiency.

\begin{algorithm}[h!]
\caption{Graph Search Algorithm}
\begin{algorithmic}[1] 
\State Landmarks $\gets \varnothing$, $g(C)\gets 0$
\State $Node_{0,0} \gets$ All predicates used in the embedding input
\State Frontier Nodes (FN) $\gets Node_{0,0}$ 
\State Frontier Nodes States(FNS) $\gets$ All unique detected states with a value of 1 in the contrastive learning algorithm

\State Negative Test ($NT$) $\gets$ Random 10 negative sample

\While {$g(c) < 1$}
\State $f(Node_{j,i})= -\frac{Number\ of\ States\ in\ Node_{j,i}}{Number\ of\ States\ in\ Node_{0,0}}-i$
\State Chosen Node ($CN_{j,i}$) $\gets$ Choose a node from softmax distribution over all $f(Node_{j,i})$ on FN
\State Node States(NS) $\gets$ Unique states with CN predicates
\For{state in NS}
    \If{(state  $\in \tau_{i,p}, \forall i$) \& ($\exists i\in NT$, state  $\notin \tau_{i,n}$)}
        \State Landmarks $\gets $ state, $g(C) \gets 1$
    \EndIf
\EndFor
\State Frontier Nodes (FN) $\gets FN/CN_{j,i}$ 
\State New Nodes($NN_{j:k+j,i+1}$) $\gets CN_{j,i}/p_k$ 
\State Frontier Nodes (FN) $\gets $ FN+ $NN_{j:k+j,i+1}$
\EndWhile
\end{algorithmic}
\end{algorithm}



\vspace{-0.9cm}
\section{Rule Generation for Attaining Landmarks using LLM}
\label{sec:typestyle}


\vspace{-0.3cm}

By employing subtask decomposition, we simplified the challenge of learning RL policy rules for a complex task by breaking it down into smaller, manageable subtasks. In this context, we employed few shot learning with the LLAMA 3.1 \cite{dubey2024llama} model to generate rules for each identified subtask. The experimental results are discussed in the following section, with details of the prompts shown in Fig. 4. The input to the LLM consists of a constant part, which includes definitions of predicates used to represent the states and general information about the environment. To create base rules, we combined the subtask with a base prompt and two rule examples from other environments, helping the model follow the rule template and grasp the logic behind the rules.

To evaluate the effectiveness of a rule, we tested the RL agent using generated template rules. If the rules fail to achieve the subtask, we refine the template rules. We record the state corresponding to the lowest reward as the failed state. Since the LLM did not generate a complete set of rules for us, we refined them by utilizing additional prompts. These prompts ask the LLM to interpret the rule and modify it by removing some predicates to increase generality or by adding predicates to enhance detail. Depending on the complexity of the subtask, we can generate rules that are either more general or more detailed.
\vspace{-0.2cm}
\begin{figure}[h!]
    \centering
    \includegraphics[width=0.6\linewidth]{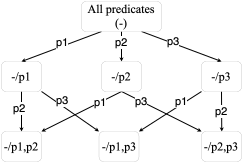}
    \caption{The schematic shows the graph generated by the algorithm for environments with three predicates. The root represents states containing all three predicates, and each subsequent level illustrates states formed by removing one predicate. Each edge indicates which predicate was removed at that node. The final leaves contain only one predicate.}
    \label{fig:example}
    
\end{figure}

\vspace{-0.5cm}
\begin{table}[h!]
    \centering
    \begin{tabular}{l|c c c}
        \toprule
        \textbf{Score} & \textbf{4 subtasks} & \textbf{3 subtasks} & \textbf{2 subtasks} \\
        \midrule
        \textbf{GetOut$^*$} & 22.86 $\pm$ 2.46  &  23.06 $\pm$ 2.37 & 23.29 $\pm$ 2.34 \\
        \textbf{GetOut}  & 22.84 $\pm$ 2.49 & 23.02 $\pm$ 2.33 & 23.31 $\pm$ 2.38 \\
        \textbf{Loot$^*$}   &  &   & 5.31 $\pm$ 0.65 \\
        \textbf{Loot}   &  &   & 5.45 $\pm$ 0.51 \\
        \bottomrule
    \end{tabular}
    \caption{Comparison of our algorithm on tasks with and without predicate knowledge, where GetOut$^*$ and Loot$^*$ exclude predicates like have-object and pickup-object, while GetOut and loot include them. Score is the agregated rewrds.}
    \label{tab:example}
\end{table} 
\section{Experiment}
\label{sec:majhead}
\vspace{-0.2cm}
The environment, adapted from the GetOut and Loot environment in \cite{delfosse2024interpretable}. GetOut has been modified to include distinct landmarks and new objects, such as two coins, a flag, and a red key. The four subtasks we refer to are: collecting two coins, collecting a flag, collecting a blue key, and then proceeding to the door.. An example state of the modified GetOut environment is shown in Fig. 6.

In Table 1, we compare the results of the algorithm in two environments: one with additional predicates and knowledge, and another with fewer predicates. We evaluate it on tasks with varying numbers of subtasks. Since we did not have labels for the landmark states in Fig. 2, we manually labeled them to evaluate the accuracy of subtask detection. Table 2 highlights the necessity of subtasks, showing results after rule generation and policy learning. Fig. 3 compares our algorithm to human generated rules, demonstrating similar success and showing that missing subtask results in task failure. Fig. 5 illustrates the comparison between the rule policy from the Nudge and a template generated rule and policy for the coin subtask.

\vspace{-0.5cm}
\section{Conclusion}
\label{sec:print}
\vspace{-0.3cm}

The paper introduces a novel method for detecting landmarks to decompose complex tasks into subtasks.
FOL state representation and leveraging LLM led us to create rule-based policies through an ILP-based RL agent. Experiments demonstrate that the algorithm is both accurate and efficient in subtask detection and that LLM-guided rule generation
This method reduces reliance on predefined logic predicates, offering a more flexible and scalable solution. Future work aims to extend the approach to real-world tasks and enhance rule fine-tuning for broader generalization.

\begin{table}[h!]
    \centering
    \begin{tabular}{l|c c c}
        \toprule
        \textbf{Score} & \textbf{4 subtasks} & \textbf{3 subtasks} & \textbf{2 subtasks} \\
        \midrule
        \textbf{GetOut$^*$/4} & 22.86 $\pm$ 2.46  &  -10.24 $\pm$ 2.05 & -14.47 $\pm$ 2.54 \\
        \textbf{GetOut$^*$/3}  &  & 23.02 $\pm$ 2.33 & -10.41 $\pm$ 2.64 \\
        \bottomrule
    \end{tabular}
    \caption{
Comparison of subtask necessity, with the x-axis showing the number of learned subtasks in our algorithm and the score representing the average return for tasks with 3 and 4 subtasks. }
    \label{tab:example}
\end{table}
\vspace{-0.8cm}

\begin{figure}[h!]
    \centering
    \includegraphics[width=0.9\linewidth]{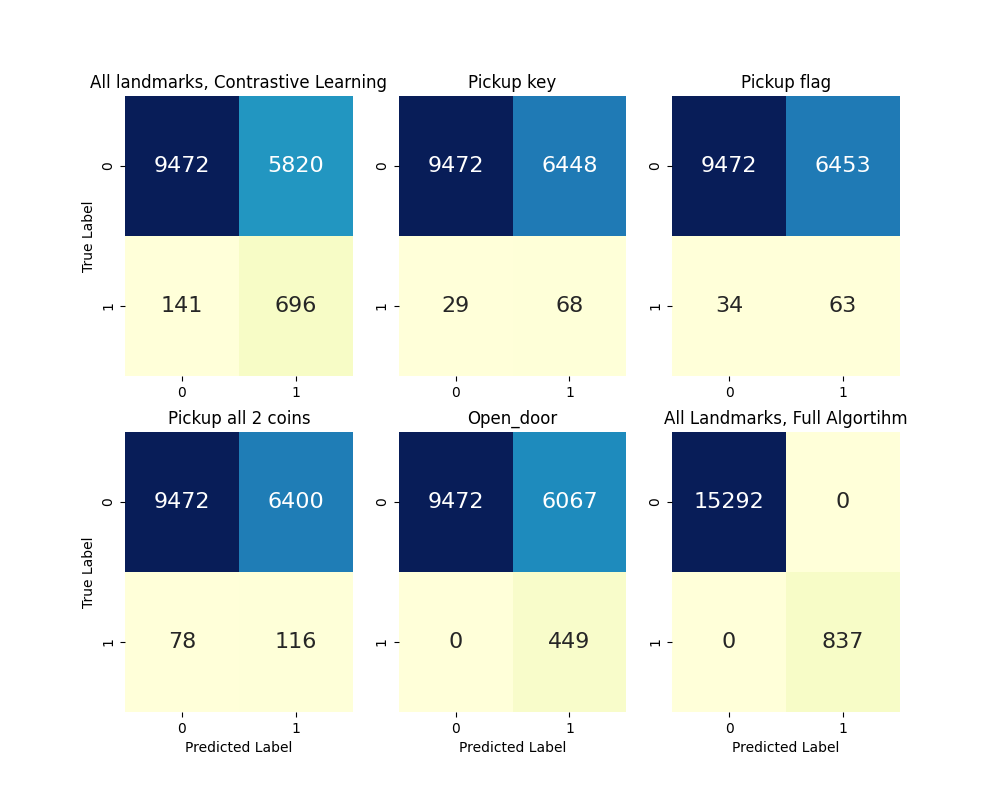}
    \caption{
Performance of landmark identification: The top-left plot shows contrastive learning results for all landmarks, and the bottom-right plot displays improvements after applying a tree graph search. Other plots focus on specific landmarks before the graph search. Recall improved from 83\% to 100\%, and precision increased from 10\% to 100\% with the search algorithm.}
    \label{fig:comparison}
\end{figure}

\vspace{-15pt}

\begin{figure}[h!]
    \centering
    \includegraphics[width=1\linewidth]{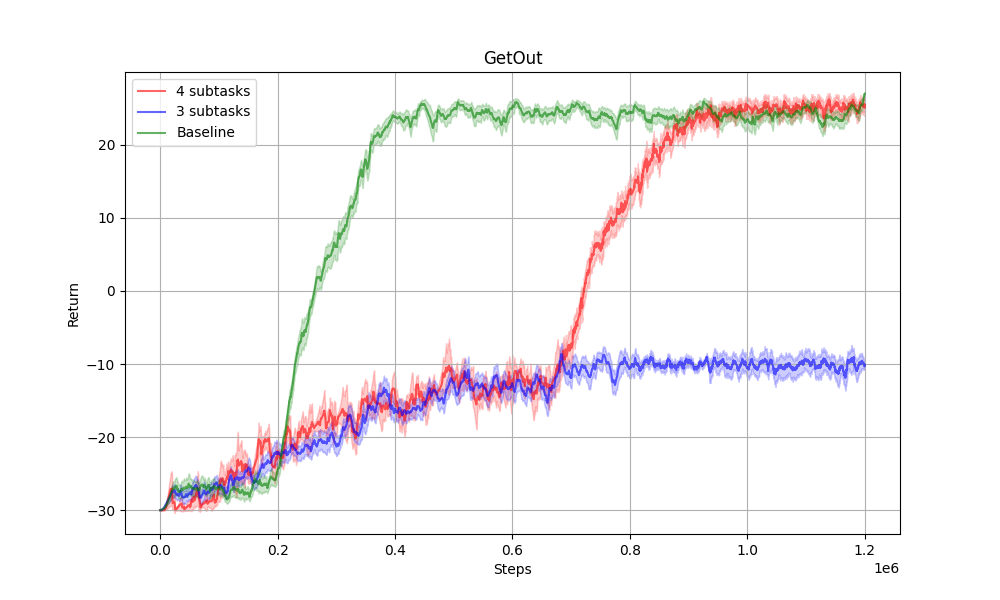}
    \caption{
Comparison of algorithm convergence: The red plot shows performance on 4 subtasks, the blue plot on 3 subtasks, and the green plot represents the ILP-RL agent using a human expert's rule template.}
    \label{fig:enter-label}
\end{figure}


\begin{figure}[h!]
    \centering
    \begin{minipage}{0.45\textwidth}
        \centering
        \includegraphics[width=1\textwidth]{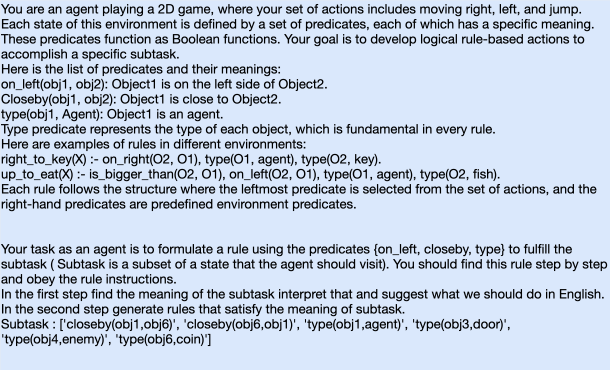}
        \label{fig:Prompt}
        \vspace{-0.4cm}
    \end{minipage}\hfill
    \begin{minipage}{0.45\textwidth}
        \centering
        \includegraphics[width=1\textwidth]{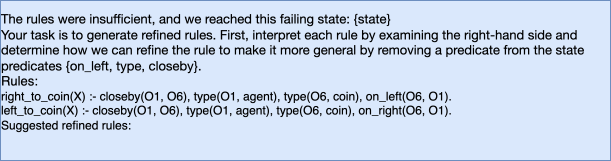}
        \caption{
Top image: Prompt for generating the base template rule, including a constant section with few shot examples from various environments and the specific coin subtask. Bottom image: Few shot learning applied to refine the template rule by generating more general rules.}
        \label{fig:Extend_prompt}
    \end{minipage}
\end{figure}

\vspace{-5cm}
\begin{figure}[h!]
    \centering
    \includegraphics[width=0.45\textwidth]{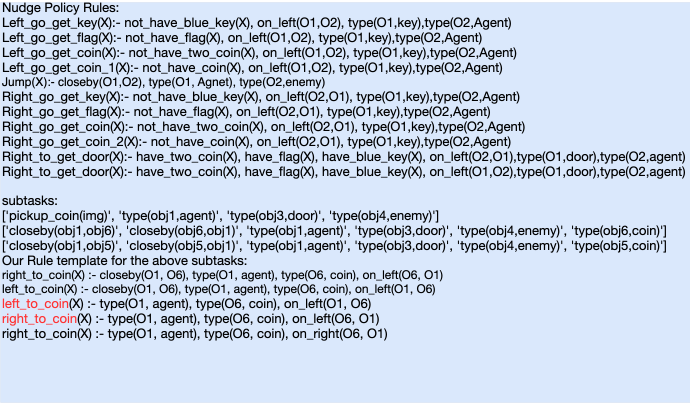}
    \caption{Comparison of the human expert's rule policy with LLM-generated rules for coin subtask. The final policy chosen by the ILP-RL agent is marked in red, demonstrating the effectiveness of subtasks in guiding smaller policy rules with less predicate or environmental information.}
    \label{fig:graph-schematic}
\end{figure}
\begin{figure}[h!]
    \centering
    \includegraphics[width=1\linewidth]{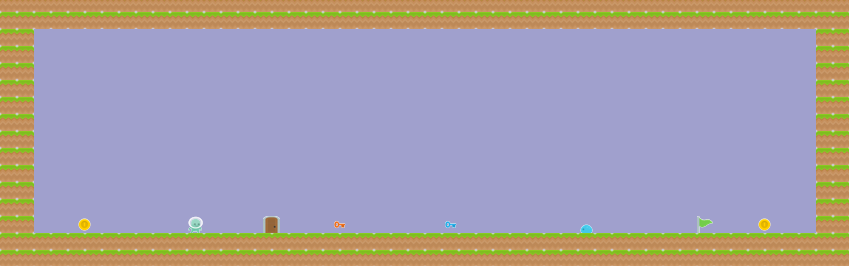}
    \caption{GetOut environment: The humanoid agent and other objects are in a defined state. The agent can move right, left, or jump.}
    \label{fig:enter-label}
\end{figure}

\vfill\pagebreak



\bibliographystyle{IEEEbib}
\bibliography{Template,refs}

\end{document}